\documentclass{article}
\usepackage{spconf,amsmath,graphicx,hyperref, amsthm, amsfonts}
\usepackage{mathtools}
\mathtoolsset{showonlyrefs}
\usepackage[table]{xcolor}
\definecolor{gradnetgreen}{RGB}{186, 247, 161}
\usepackage{enumitem}

\def\x{{\mathbf x}}

\def\W{{\mathbf W}}
\def\y{{\mathbf y}}

\providecommand{\x}{\mathbf{x}}

\providecommand{\z}{\mathbf{z}}
\providecommand{\W}{\mathbf{W}}

\renewcommand{\b}{\mathbf{b}}

\providecommand{\R}{\mathbb{R}}

\newtheorem{theorem}{Theorem}
\newtheorem{lemma}{Lemma}

\newtheorem{definition}{Definition}
\theoremstyle{remark}

\title{Learning Gradients of Homogeneous Functions}
\name{Mayur S. Gowda*, Srinivasa Pranav*, Shreyas Chaudhari*, Jos\'e M. F. Moura \thanks{*Equal contribution.\\
Authors partially supported by NSF Grants CNS-2106891, CNS-2409138, and CCF-2327905. S. Chaudhari and S. Pranav partially supported by NSF Graduate Research Fellowships (GRFP; Grants DGE1745016, DGE2140739). S. Pranav partially supported by an ARCS Fellowship.}}
\address{Department of Electrical and Computer Engineering, Carnegie Mellon University}
\begin{document}
\ninept
\maketitle
\begin{abstract}
Homogeneous gradient functions have widespread significance, including applications in maximum energy subspace selection and retrieval problems.
Euler's homogeneous function theorem specifies how the potential associated with a positively $k$-homogeneous gradient field can be recovered directly from the field itself. We use Euler's theorem to build on GradNets~\cite{GRADNETSTSP} and propose a homogeneous gradient network architecture that is guaranteed to correspond to the gradient of a homogeneous function. 
Then, we describe how monotone homogeneous GradNets universally approximate the gradients of $k$-homogeneous convex functions.
The inductive bias of homogeneous GradNets enables them to outperform baseline ICNNs by up to 7dB in our max energy subspace projection experiment and by up to 13\% higher accuracy, 98\% fewer parameters, and 70\% fewer FLOPS in our amortized inner product search experiment.
\end{abstract}

\section{Introduction}
\label{sec:intro}
Many signal processing and machine learning problems have structured solutions that leverage prior knowledge about the problem. Vector fields for physics and optimization problems often obey common properties like scaling consistently with the input (positive homogeneity) and corresponding to the gradient of an underlying scalar potential (conservativity). Such requirements arise in learned convex regularizers~\cite{mukherjee2024data, goujon2023neural}, optimal transport \cite{brenier1991polar, CAMSAPGradNets}, and in retrieval problems where the optimal map is the gradient of a convex support function~\cite{AMIPS}.
Specifically, controlling the \emph{degree of positive homogeneity} of the underlying potential is an important design requirement in applications where the target field has known scaling behavior.

\textbf{Maximum inner product search (MIPS):} The support function of a
fixed database, $\sigma_{\mathcal Y}(\x)=\max_i \langle \x, \y_i\rangle$,
is convex and positively $1$-homogeneous, and its gradient at $\x$ is
exactly the maximizing database vector. The runtime scales linearly in the dimension and the database size; amortizing MIPS requires
learning the gradient of a $1$-homogeneous convex function~\cite{AMIPS}.
\textbf{Maximum-energy subspace selection and projection in array and radar applications:} candidate signal or interference subspaces can be compared by the energy they capture from a received snapshot $\x$ ~\cite{MatchedSubspaceDetector, orlando2022unified}. For a candidate subspace with projection matrix $P_i\succeq0$, this energy is
$
\|P_i\x\|_2^2=\x^\top P_i\x.
$
The resulting pointwise maximum of quadratic forms is convex and positively $2$-homogeneous. Efficient amortized identification of dominant signal or interference subspaces is useful in high-throughput radar processing.
\textbf{Convex regularizers} used in inverse problems (e.g., norms, gauge functions, and total variation-type penalties) are positively 
$1$-homogeneous by design, as the scaling enables consistent behavior across signal amplitudes~\cite{benning2018modern}; computing their proximal or gradient steps requires exactly the structure we study in this paper. Global optimization guarantees have also been prove using positive homogeneity~\cite{GlobalOptimality}.

Neural network architectures have been proposed to approximate gradient functions, but they do not explicitly parameterize the degree of positive homogeneity of the underlying potential. For examples, input convex neural networks~\cite{ICNN} guarantee convexity of the potential by construction, but leave the scaling behavior of the learned field unconstrained. A natural solution is to use an unconstrained (or convexity-constrained) neural network and encourage homogeneity through regularization or a wrapper function. This strategy struggles to simultaneously guarantee that the learned field is integrable and corresponds to a convex homogeneous function.

In this work, we encode prior knowledge into the neural network architecture to guarantee homogeneity, conservativity, and monotonicity. By enforcing the correct functional properties in neural network architectures, we guarantee that every parameter setting yields a valid gradient field with the desired structure. Instead of first parameterizing the potential function, we directly parameterize the gradient function with a modular gradient network (GradNet-M) that we proposed in~\cite{GRADNETSTSP}. Crucially, we leverage Euler's homogeneous function theorem and use homogeneous module activations to express each module's potential as an inner product.

\textbf{Contributions:} Our contributions are threefold. (1) We propose a gradient network architecture~\cite{GRADNETSTSP} whose output is guaranteed to be the gradient of a positively $k$-homogeneous (convex) function. (2) We prove that the proposed network universally approximates (sub)gradients of positively $k$-homogeneous convex functions. (3) We evaluate on two tasks with different homogeneity degrees: amortized maximum quadratic energy subspace selection ($k=2$) and amortized maximum inner product search ($k=1$), comparing against unconstrained and convexity-constrained baselines.

\section{Related Work}
\label{sec:related_work}





\textbf{Convex neural networks.}
Input convex neural networks (ICNNs\cite{ICNN}) parameterize a scalar potential and require nonnegative weights to ensure that the output is convex with respect to the input. Computing the gradient field requires differentiating the neural network via backpropagation at inference time.
To address ICNNs' training difficulties caused by nonnegative weights, input convex gradient networks (ICGNs\cite{ICGN}) parameterizes the Hessian and recovers the gradient via numerical integration. Neither of these architectures model the gradient directly nor constrain the degree of positive homogeneity of the underlying potential. \cite{AMIPS} applies a homogenization wrapper $\mathcal{H}[g](\x)=\|\x\|\,g(\x/\|\x\|)$ to make the ICNN (called SupportNet) match the exact $1$-homogeneity of the true MIPS support function. However, the wrapped network may no longer correspond to the gradient of a convex function. Counterexample: $g(\x)=\x_1^2$; the Hessian of $\mathcal{H}[g]$ at $\x = [1,0,\dots,0]^\top$ has an eigenvalue of $-1$. Our proposed homogeneous GradNet constructively enforces the target homogeneity degree in the parameterization and ensures convexity and homogeneity hold simultaneously (see Sec.~\ref{sec:method}).

\textbf{Gradient Networks} (GradNets~\cite{GRADNETSTSP}) directly parameterize the gradient of a convex potential, avoiding a separate potential network. They have been used in applications ranging from optimal transport~\cite{CAMSAPGradNets} and generative flows~\cite{GRADNETSICASSP, noufal2026generative} to physics-informed safe control~\cite{jacob2025generalizable}, electromagnetic fields~\cite{Hinkkanen2026gradient}, and material science~\cite{jihong2026monotone}.

\textbf{Subspace detection and selection} is a long-studied problem in array and radar processing.
Comparing an observation against multiple candidate signal or
interference subspaces arises in matched subspace detection~\cite{MatchedSubspaceDetector}, adaptive
beamforming~\cite{BeamForming}, and related detection-theoretic formulations where a
statistic such as $\x^\top P_i \x$ is computed for each candidate
subspace projector $P_i$ and thresholded or maximized over
$i$.
These classical formulations typically assume a small, fixed number of
candidate subspaces evaluated exactly at each observation. Our first
experiment (Sec.~\ref{sec:max_energy}) instead treats the maximum
projected energy $F(\x)=\max_i \x^\top P_i\x$ as the learning objective since $F$ is convex and positively $2$-homogeneous, and its
gradient at points of unique maximization is exactly $2P_{i^*}\x$.

\textbf{Maximum inner product and subspace search} require an exhaustive $\mathcal{O}(Nd)$ comparison of each query against a database, which becomes prohibitive at scale.
Classical approaches instead build search structures such as locality-sensitive hashing~\cite{AsymmetricALSH} or tree- and graph-based indices~\cite{TreeCones}. %
Alternatively, \cite{AMIPS} proposes amortizing the search by learning a direct query-to-answer map using the gradient of the database support
function. In Sec.~\ref{sec:amips}, we adopt their MIPS setup and compare to their proposed methods.

\textbf{Positively homogeneous convex functions and regularizers} are classically studied as gauge
functions and sublinear functions~\cite{rockafellar1970convex} and used
throughout inverse problems, where the scaling behavior of the
regularizer makes it behave consistently across signal
amplitudes.
Our universal approximation results (Sec.~\ref{sec:theory}) build on the density of pointwise maxima
of supporting hyperplanes~\cite{ConvexPotentialFlows}; the
extension from $1$-homogeneous to $k$-homogeneous
convex functions uses the factorization result of~\cite{huang2020generalizing}.
 
\section{Problem Statement}
\label{sec:problem_statement}

We consider the problem of learning vector fields that are gradients of positively homogeneous (convex) functions. Such structure arises in applications where the learned mapping must simultaneously satisfy integrability, scale-consistency, and convexity constraints. Rather than learning an unconstrained vector-valued function and subsequently enforcing these properties, we leverage the prior knowledge about the optimal solution to construct a neural network architecture whose output satisfies the properties by construction.

\textbf{Objective}: construct a parameterized vector field $f_\theta:\mathbb{R}^d\rightarrow\mathbb{R}^d$
such that, for every choice of the learnable parameters $\theta$, there exists a potential $F_\theta$ satisfying $f_\theta(\x)=\nabla F_\theta(\x)$ where $F_\theta$ is positively $k$-homogeneous. In addition, we seek a parameterization that permits the homogeneity degree of the potential to be controlled explicitly.

\begin{definition}[Positive $k$-Homogeneity]
Let $k\in \mathbb{N}$. A function $f$ is positively $k$-homogeneous if $f(t\x) = t^kf(\x)$ for every $t \geq 0$.
\end{definition}

Positive homogeneity imposes a simple scaling relationship between a function and its values along rays from the origin. An important consequence is that the gradient of a differentiable positively $k$-homogeneous function is positively $(k-1)$-homogeneous:
\begin{equation}
\nabla F(t\x)=t^{k-1}\nabla F(\x)
\end{equation}
Thus, if the target vector field is the gradient of a positively $k$-homogeneous potential, its degree of homogeneity is determined by that of the underlying potential. \textbf{Euler's homogeneous function theorem} provides the following pointwise characterization that enables us to recover the potential from its gradient: for a (sub)differentiable positively $k$-homogeneous function,
\begin{align}
\x^\top\nabla F(\x) =kF(\x) \implies 
F(\x) =\frac{1}{k}\x^\top\nabla F(\x)
\end{align}
This identity allows us to express the potential associated with a learned homogeneous gradient field directly in terms of an inner product with the learned field, without requiring a separate neural network to represent the potential.
Next, we leverage these properties to construct the proposed homogeneous gradient network.
 
\section{Homogeneous Gradient Networks}
\label{sec:method}
We propose a neural network architecture that is, by construction, guaranteed to correspond to the gradient of a positively $nk$-homogeneous function:
\begin{align}
    f(\x) &= \sum_{i=1}^{M} \alpha_i\, k \big(\x^\top g_i(\x)\big)^{k-1} g_i(\x) \label{eq:arch-f}\\
    g_i(\x) &= \W_i^\top \sigma(\W_i \x) \label{eq:arch-g}
\end{align}
The architecture has $M$ learnable modules and resembles the modular gradient networks (GradNet-M) that we previously proposed in~\cite{GRADNETSTSP,GRADNETSICASSP}.
The hyperparameter $k$ is user-defined and controls the potential function $F$'s degree of homogeneity. The activation $\sigma$ is constrained to be the gradient of $\phi$, a positively $n$-homogeneous scalar-valued function. Each module has unconstrained learnable parameters: scalar mixing coefficients $\alpha_i$ and weight matrices $\W_i$.

The neural network corresponds to the \emph{monotone} gradient of a \emph{convex} positively $nk$-homogeneous function if the following additional requirements are satisfied: the coefficients $\alpha_i$ are constrained to be nonnegative and the $\phi$ are additionally convex and nonnegative.
As we prove next, $f$ corresponds to the gradient of a (convex) positively $kn$-homogeneous function $F$ under these conditions.

\begin{theorem}
    Let $\phi: \R^d \to \R$ be positively $n$-homogeneous and $\sigma: \R^d \to \R^d$ be the gradient of $\phi$. Then neural networks of the form~\eqref{eq:arch-f} are guaranteed to correspond to gradients of positively $kn$-homogeneous functions. If every $\alpha_i \geq 0$ and $\phi$ is additionally convex and nonnegative, then $f$ is guaranteed to correspond to the \emph{monotone} gradient of a \emph{convex} positively $kn$-homogeneous function. 
\end{theorem}
\begin{proof}
Let $\phi_i=\phi\circ\W_i$ so that $g_i(\x)=\nabla\phi_i(\x)$. By Euler's identity, $\x^\top g_i(\x) = n\phi_i(\x)$. Substituting into \eqref{eq:arch-f} gives
\begin{equation}
    f(\x) \!=\! n^{k-1}\!\sum_i \alpha_i k\,\phi_i(\x)^{k-1}\nabla\phi_i(\x)\ \!=\! n^{k-1}\!\sum_i \alpha_i \nabla\left(\phi_i(\x)^k\right)    
\end{equation}
Therefore, $f(\x)=\nabla F(\x)$ with 
$F(\x)=n^{k-1}\sum_i\alpha_i\phi_i(\x)^k$.
Each $\phi_i$ is positively $n$-homogeneous, so $\phi_i^k$ is positively
$kn$-homogeneous and $F$ is positively $kn$-homogeneous.

If $\phi$ is additionally convex and nonnegative, then each $\phi_i$ is convex and nonnegative. For $k\geq1$ and $\alpha_i\geq0$,
$\alpha_i\phi_i^k$ is convex because $t\mapsto t^k$ is convex and
nondecreasing on $t\geq0$.
Therefore, $F$ is convex and hence $f=\nabla F$
is the gradient of a convex, positively $kn$-homogeneous function.
\end{proof}

The key observation is that, when $\sigma$ is homogeneous, Euler's identity converts the scalar quantity $\x^\top g_i(\x)$ into the potential associated with the module. This permits nonlinear powers of the potential to be differentiated analytically, yielding a gradient field while preserving positive homogeneity. The theorem also provides an analytic form for the antiderivative of a homogeneous function.
Each module \eqref{eq:arch-g} is a bias-free mGradNet-M module of \cite{GRADNETSTSP} whose modular architecture takes the form $\sum_i s_i(\z_i)\W_i^\top\sigma_i(\z_i)$ with $\z_i = \W_i\x + \b_i$, $\sigma_i = \nabla s_i$, and $s_i$ convex and nonnegative; such a module is $\nabla[\tfrac{1}{2}s_i^2]$. Two consequences of homogeneity distinguish the proposed form: there is no separate evaluation of $s_i$, and the non-negativity requirement on $s_i$ is automatically satisfied.

\section{Universal Approximation Results}
\label{sec:theory}

In this section, we prove universal approximation results for the homogeneous monotone gradient network architecture. First we consider approximators of convex $1$-homogeneous potential functions.

\begin{lemma}
    Pointwise maximum of linear functions is dense in the space of convex positively 1-homogenous functions.
    \label{lemma:max_of_linear}
\end{lemma}
\begin{proof}
    This follows from Proposition 2 of~\cite{ConvexPotentialFlows}, which shows that the pointwise maximum of supporting hyperplanes $L_{\y}(\x) = \nabla F(\y)^\top(\x - \y) + F(\y)$ can uniformly approximate $F$ (where $\nabla F(\y)$ is a subgradient of $F$ evaluated at $\y$). If $F$ is homogeneous, then by Euler's identity $$L_{\y}(\x) = \nabla F(\y)^\top(\x - \y) + F(\y) = \nabla F(\y)^\top \x$$ and the supporting hyperplanes are linear instead of affine.
\end{proof}

We show that a subgradient of the universal approximators in extend Lemma~\ref{lemma:max_of_linear} yield universal approximators of gradients of convex $1$-homogeneous functions.

\begin{theorem}
    The function $g(\x)$ in \eqref{eq:arch-g} is a universal approximator of monotone (sub)gradients of convex positively 1-homogeneous functions when the activation $\sigma$ is argmax. \label{thm:1-homogeneous}
\end{theorem}
\begin{proof}
    Let $F$ be a 1-homogeneous convex function. By Lemma~\ref{lemma:max_of_linear}, there exists a sequence of convex approximators (i.e., maximum of linear functions) that converges pointwise to $F$. By Theorem 24.5 in \cite{rockafellar1970convex}, a subsequence of subdifferentials of these approximators becomes increasingly similar to those of $F$. When $g(\x)$ in \eqref{eq:arch-g} has argmax activation, it is precisely a subgradient of the maximum of linear functions.
\end{proof}

Next, we consider the broader class of convex positively $k$-homogeneous functions and express it in terms of convex positively $1$-homogeneous functions.  will lead to extensions of.

\begin{lemma}
    Let $F$ be convex and positively $k$-homogeneous with $k > 1$. Then $F$ is non-negative and there exists convex positively 1-homogeneous $H$ such that $F(\x) = H(\x)^k$. \label{lemma:k_in_terms_of_1}
\end{lemma}
\begin{proof}
    Let $t \in (0,1)$. Homogeneity of $F$ implies $F(\mathbf{0}) = 0$. Convexity of $F$ implies $(t^k - t)F(\x) \leq 0$ because
    \begin{equation}
        t^k F(\x) = F(t\x + (1-t)\mathbf{0}) \leq tF(\x) + (1-t)F(\mathbf{0}) = tF(\x)
    \end{equation}
    Since $t \in (0,1)$ and $k > 1$, this implies $F(\x) \geq 0$. The remainder of the proof follows from Lemma D.1 in~\cite{huang2020generalizing}.
\end{proof}

Next, we leverage Lemma~\ref{lemma:k_in_terms_of_1} to show universal approximation of monotone gradients of convex positively $k$-homogeneous functions.

\begin{theorem}
    The function $f(\x)$ in \eqref{eq:arch-f} is a universal approximator of monotone (sub)gradients of convex positively $k$-homogeneous functions, where $k > 1$, when the activation $\sigma$ is argmax.
\end{theorem}
\begin{proof}
    Let $F$ be convex positively $k$-homogeneous, with $k > 1$. By Lemma~\ref{lemma:k_in_terms_of_1}, there exists convex positively $1$-homogeneous $H$ such that $F(\x)=H(\x)^k$. By Lemma~\ref{lemma:max_of_linear}, there exists a sequence of maximum of linear functions that converges to $H$ pointwise. Since composing with a continuous function preserves pointwise convergence, the sequence raised to the power $k$ also converges pointwise to $H(\x)^k$, and the result follows by Theorem 24.5 in~\cite{rockafellar1970convex}.
\end{proof}

\section{Experimental Evaluation}
\label{sec:experiments}

\subsection{Amortized Max Energy Subspace}
\label{sec:max_energy}

Subspace projection is a fundamental operation in signal processing that has applications including matched-subspace detection and hyperspectral target detection, where observations are compared with low-dimensional signal or target subspaces. 
We consider $K$ candidate $r$-dimensional subspaces $\{\mathcal{S}_i\}_{i=1}^K$ in $\mathbb{R}^d$. Let $U_i\in\mathbb{R}^{d\times r}$ contain an orthonormal basis for $\mathcal{S}_i$ and $P_i = U_i U_i^\top$ be the corresponding orthogonal projector. For an observation $x\in\mathbb{R}^d$, the energy captured by the $i$th subspace is $\|P_i x\|_2^2 = x^\top P_i x$. We study the maximum subspace energy projection problem: project onto the candidate subspace that captures the most signal energy
\begin{equation}
    F(x) = \max_{i\in[K]} x^\top P_i x.
    \label{eq:max_energy}
\end{equation}
When the maximizing subspace is unique, $F$ is differentiable and
$\nabla F(x) = 2P_{i^*}x \label{eq:max_energy_gradient}$. We use this to evaluate the accuracy of the projection amortization.
We construct $K=10$ candidate subspaces and vary the ambient dimension as $d\in\{16,32,64,128,256\}$ with subspace dimension $r=d/4$. Observations are generated from randomly selected latent subspaces with additive isotropic Gaussian noise ($\sigma = 0.5$), yielding samples that concentrate around the candidate subspaces while allowing overlap. We respectively construct 100K, 10K, and 10K samples for training, validation, and testing, and evaluate all $K$ projected energies to construct the targets.

\begin{figure}[t]
    \centering
    \includegraphics[width=\columnwidth]{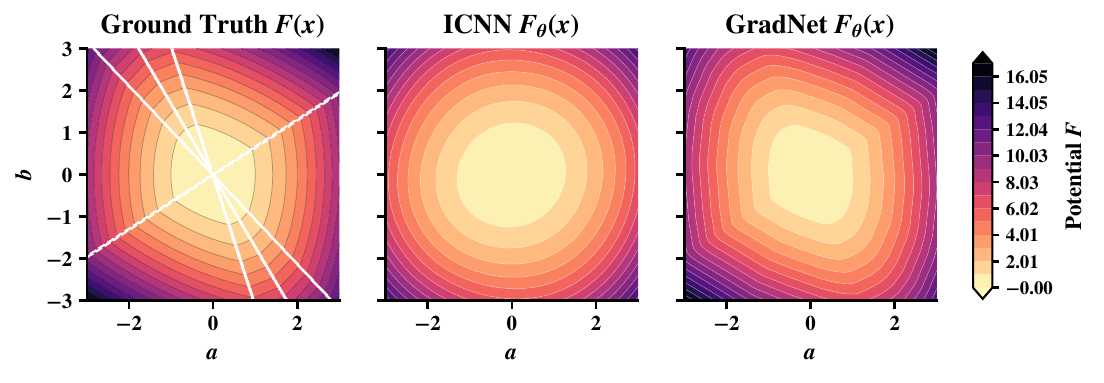}
    \caption{Ground-truth and learned potentials on a 2-D orthonormal
    slice of a $d = 16$ maximum energy subspace problem.
    White lines indicate boundaries where the maximizing projector
    changes.}
    \label{fig:potential_slice}
\end{figure}

Figure~\ref{fig:potential_slice} visualizes the target and learned
potentials on the slice $x=av_1+bv_2$. The ground-truth potential is
piecewise quadratic, with boundaries corresponding to changes in the
maximizing projector. GradNet more closely reproduces this geometry than ICNN.
We compare an input convex neural network (ICNN)~\cite{ICNN} baseline with our proposed $1$-homogeneous GradNet architecture ($k=2,n=1$) using weighted softmax activation. Let $g_i(x;\tau_t)$ denote the output of the $i$th GradNet module. The predicted field is
\begin{equation}
    g_\theta(x;t)
    =
    \sum_{i=1}^{M}
    \alpha_i
    \left[
        2\,\mathrm{ReLU}
        \left(
            x^\top g_i(x;\tau_t)
        \right)
        g_i(x;\tau_t)
    \right],
    \label{eq:proposed_gradnet}
\end{equation}
where $\alpha_i>0$ are learned mixture coefficients and $M=4$. During training, the softmax logit scale $\tau_t$ is increased linearly from $0.25$ to $100$, progressively sharpening the module responses. We scale both architectures' hidden dimensions such that each model has $1024d$ parameters. Each model is trained for $100$ epochs
using Adam and MSE loss $
    \mathcal{L}_{\mathrm{grad}}
    =
    \frac{1}{Nd}
    \sum_{n=1}^{N}
    \left\|
        g_\theta(x_n)-g^*(x_n)
    \right\|_2^2.
    \label{eq:gradient_mse}
$
We evaluate the lowest validation MSE checkpoint on the held-out test set. 

\begin{figure}[t]
    \centering
    \includegraphics[width=\columnwidth]{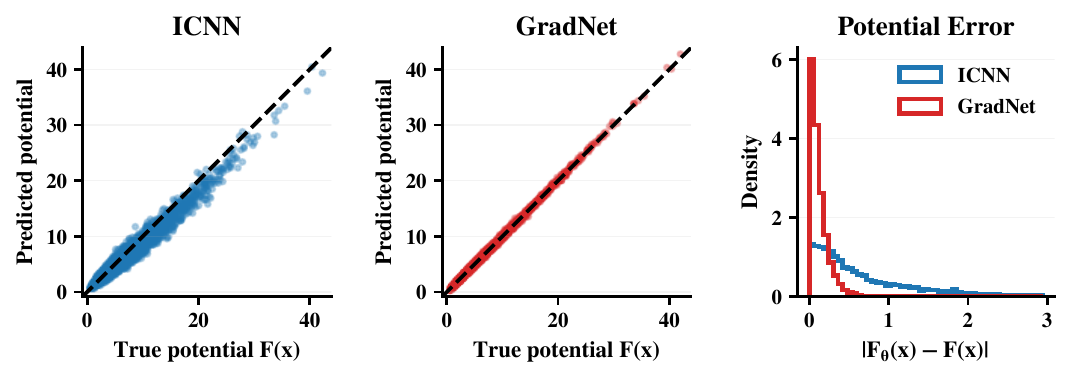}
    \caption{$d=16$ maximum energy subspace problem.
    Left and center: true vs. predicted potentials for ICNN and
    GradNet; dashed line denotes exact agreement. Right:
    absolute potential error distributions.}
    \label{fig:potential_full16d}
\end{figure}

Figure~\ref{fig:potential_full16d} evaluates the learned potentials
on the complete $d=16$ test set. GradNet predictions closely follow
the ground-truth diagonal, with absolute errors more concentrated
near zero than those of the ICNN.
Across all dimensions, Table.~\ref{tab:max_energy_mse} shows that explicitly incorporating the $1$-homogeneity of the target gradient into our GradNet architecture enables it to achieve significantly lower test gradient MSE than the ICNN baseline -- with up to $7$dB improvement.

\begin{table}[!ht]
\centering
\caption{Maximum projected
subspace energy task test gradient MSE (dB) across ambient dimensions $d$. Lower is better.}
\label{tab:max_energy_mse}
\setlength{\tabcolsep}{3.5pt}
\renewcommand{\arraystretch}{1.05}
\scriptsize
\begin{tabular}{c|ccccc}
\hline
$d$ & 16 & 32 & 64 & 128 & 256 \\
\hline
ICNN
& -5.04 & -4.36 & -4.13 & -3.99 & -3.89 \\
GradNet
& \textbf{-11.97} & \textbf{-9.74} & \textbf{-7.66}
& \textbf{-5.63} & \textbf{-4.89} \\
\hline
\end{tabular}
\end{table}









\subsection{Amortized Max Inner Product Search}
\label{sec:amips}

We consider queries $x\in\mathbb{R}^{d}$, keys $y_i\in\mathbb{R}^{d}$, a fixed database $\mathcal{Y}=\{y_i\}_{i=1}^{N}$ and the maximum inner product search (MIPS) problem:
\begin{align}
    i^*(x)
=
\arg\max_{i\in[N]} \langle x,y_i\rangle
\end{align}
with corresponding maximizing key
$y^*(x)=y_{i^*(x)}$.
When using a fixed database, repeatedly evaluating all $N$ inner products incurs $\mathcal{O}(Nd)$ cost per query. Amortized MIPS \cite{AMIPS} instead exploits the fixed database to learn a query-dependent map that approximates the maximizing key, shifting computation to an offline training stage. We follow the setup in~\cite{AMIPS} and define the database support function as the following convex positively $1$-homogeneous function: $
\sigma_{\mathcal{Y}}(x)
=
\max_{i\in[N]}\langle x,y_i\rangle$.
When the maximizing key is unique, the support function's gradient is exactly the maximizing database vector.
We exploit this observation by learning a parameterized gradient field
$g_{\theta}:\mathbb{R}^{d}\rightarrow\mathbb{R}^{d}$ satisfying $
g_{\theta}(x)
\approx
\nabla\sigma_{\mathcal{Y}}(x)
=
y^*(x).
$
Thus, the learned model maps a query directly to an approximation of
the maximizing database vector. We perform exact exhaustive retrieval to obtain training and test data.
Following the setup of~\cite{AMIPS}, we evaluate on two text-retrieval datasets: FIQA and Quora. In the restricted setting, training uses only the original query embeddings. In the noise-augmented setting, each query is augmented with 10 noisy copies while keeping the
database fixed.
During training, we optimize the sum of a target-vector objective $\mathcal{L}_{\mathrm{target}}
=
\mathbb{E}_{x}
\left[
\left\|
g_{\theta}(x)-y^*(x)
\right\|_2^2
\right]$ and a support-value consistency objective $\mathcal{L}_{\mathrm{score}}
=
\mathbb{E}_{x}
\left[
\left(
\langle x,g_{\theta}(x)\rangle
-
\sigma_{\mathcal{Y}}(x)
\right)^2
\right]$.

We parameterize
$g_{\theta}$ as the following monotone gradient network
\begin{equation}
  g_{\theta}(x)
=
\sum_{m=1}^{M}
\lambda_m W_m^{\top}
\left[
a\,\mathrm{softmax}(\beta W_m x)
-
b\,\mathrm{softmax}(-\beta W_m x)
\right],  
\end{equation}

where $\lambda_m>0$, $a,b>0$, and $\beta>0$. The positive mixture
coefficients are parameterized as
$\lambda_m=\mathrm{softplus}(\alpha_m)$.

{
\setlength{\textfloatsep}{4pt}
\begin{table}[t]
\centering
\caption{Retrieval performance and efficiency under the restricted
(no-noise) setting (25K training steps).}
\label{tab:restricted_results}
\scriptsize
\setlength{\tabcolsep}{2.2pt}
\renewcommand{\arraystretch}{1.02}


\begin{tabular}{lccccc}

\hline
Model &
Acc. $\uparrow$ &
MRR $\uparrow$ &
R@0.01\% $\uparrow$ &
Params $\downarrow$ &
Proj. FLOPs $\downarrow$ \\
\hline

\multicolumn{6}{c}{\textbf{FIQA}} \\
\hline
SupportNet-M4 \cite{AMIPS}
& 52.70 & 0.6545 & 81.46 & 2.15M & 73.2K \\

KeyNet-M4 \cite{AMIPS}
& 51.52 & 0.6434 & 80.26 & 2.30M & 29.8K \\

\rowcolor{gradnetgreen}
GradNet 144-M8
& 63.71 & 0.7553 & 90.83 & 442K & \textbf{28.5K} \\

\rowcolor{gradnetgreen}
GradNet 128-M9
& 63.87 & 0.7562 & \textbf{91.14} & 442K & 29.3K \\

\rowcolor{gradnetgreen}
GradNet 112-M10
& \textbf{64.25} & \textbf{0.7591} & 91.06 &
\textbf{430K} & 29.3K \\

\hline
\multicolumn{6}{c}{\textbf{Quora}} \\
\hline

SupportNet-M4 \cite{AMIPS}
& 88.97 & 0.9311 & 99.84 & 20.03M & 224.2K \\

KeyNet-M4 \cite{AMIPS}
& 88.19 & 0.9255 & 99.79 & 20.20M & 100.9K \\

\rowcolor{gradnetgreen}
GradNet 144-M8
& 92.00 & 0.9540 & \textbf{99.99} & 442K &
\textbf{28.5K} \\

\rowcolor{gradnetgreen}
GradNet 128-M9
& 92.03 & 0.9542 & \textbf{99.99} & 442K & 29.3K \\

\rowcolor{gradnetgreen}
GradNet 112-M10
& \textbf{92.12} & \textbf{0.9544} & \textbf{99.99} &
\textbf{430K} & 29.3K \\

\hline
\end{tabular}%
\end{table}
}

{
\setlength{\textfloatsep}{4pt}
\begin{table}[t]
\centering
\caption{Retrieval performance and efficiency with noise-augmented
training using 10 noisy samples per query (10K training steps).}
\label{tab:noise10_results}
\scriptsize
\setlength{\tabcolsep}{2.2pt}
\renewcommand{\arraystretch}{1.02}

\begin{tabular}{lccccc}
\hline
Model &
Acc. $\uparrow$ &
MRR $\uparrow$ &
R@0.01\% $\uparrow$ &
Params $\downarrow$ &
Proj. FLOPs $\downarrow$ \\
\hline

\multicolumn{6}{c}{\textbf{FIQA}} \\
\hline

SupportNet-M4 \cite{AMIPS}
& 51.76 & 0.6449 & 80.53 & 2.15M & 73.2K \\

KeyNet-M4 \cite{AMIPS}
& 63.20 & 0.7354 & 86.60 & 2.30M & 29.8K \\

\rowcolor{gradnetgreen}
GradNet 144-M8
& 64.50 & \textbf{0.7618} & \textbf{91.52} &
442K & \textbf{28.5K} \\

\rowcolor{gradnetgreen}
GradNet 128-M9
& 64.19 & 0.7592 & 91.43 & 442K & 29.3K \\

\rowcolor{gradnetgreen}
GradNet 112-M10
& \textbf{64.60} & 0.7614 & 91.28 &
\textbf{430K} & 29.3K \\

\hline
\multicolumn{6}{c}{\textbf{Quora}} \\
\hline

SupportNet-M4 \cite{AMIPS}
& 87.77 & 0.9242 & 99.89 & 20.03M & 224.2K \\

KeyNet-M4 \cite{AMIPS}
& 87.82 & 0.9220 & 99.77 & 20.20M & 100.9K \\

\rowcolor{gradnetgreen}
GradNet 144-M8
& 91.54 & 0.9511 & \textbf{99.99} &
442K & \textbf{28.5K} \\

\rowcolor{gradnetgreen}
GradNet 128-M9
& \textbf{91.63} & \textbf{0.9516} & \textbf{99.99} &
442K & 29.3K \\

\rowcolor{gradnetgreen}
GradNet 112-M10
& \textbf{91.63} & 0.9515 & 99.98 &
\textbf{430K} & 29.3K \\

\hline
\end{tabular}%
\end{table}
}

Tables~\ref{tab:restricted_results} and~\ref{tab:noise10_results}
compare GradNet models with SupportNet and KeyNet under
restricted and noise-augmented training settings, respectively.
In addition to retrieval accuracy, we report mean reciprocal rank (MRR), Recall@0.01\%, number of trainable parameters, and
projection FLOPs per query. Across both datasets and training settings, GradNet outperforms both baselines from~\cite{AMIPS}. On FIQA, GradNet improves top-1 accuracy, MRR, and Recall@0.01\% over both baselines while using $\approx80\%$ fewer parameters. On Quora, GradNet similarly outperforms both baselines across
the retrieval metrics while requiring roughly $98\%$ fewer parameters and $70\%$ fewer projection FLOPs. Furthermore, under comparable parameter and projection-FLOP budget, GradNet performance is robust to the particular choice of architectural hyperparameters.

\section{Conclusion}
We introduced homogeneous GradNets, which guarantee positive homogeneity and conservativity by construction, and convexity under mild additional constraints. We proved universal approximation for gradients of convex, positively homogeneous functions. On max energy subspace selection, GradNet cuts test gradient MSE by up to 7dB over ICNNs; on amortized retrieval, it outperforms baselines across all metrics while using up to 98\% fewer parameters and 70\% fewer projection FLOPs. These results demonstrate benefits of encoding homogeneity directly into the architecture.

\newpage
\bibliographystyle{IEEEbib}
\bibliography{strings,refs}

@article{huang2020generalizing,
  title={{Generalizing Lieb's Concavity Theorem via Operator Interpolation}},
  author={Huang, De},
  journal={Advances in Mathematics},
  volume={369},
  pages={107208},
  year={2020},
  publisher={Elsevier}
}

@article{GRADNETSTSP,
  title={{Gradient Networks}},
  author={Chaudhari, Shreyas and Pranav, Srinivasa and Moura, Jos{\'e} M. F.},
  journal={IEEE Transactions on Signal Processing},
  volume={73},
  pages={324--339},
  year={2024},
  publisher={IEEE}
}

@INPROCEEDINGS{GRADNETSICASSP,
  author={Chaudhari, Shreyas and Pranav, Srinivasa and Moura, José M. F.},
  booktitle={ICASSP 2023 - 2023 IEEE International Conference on Acoustics, Speech and Signal Processing (ICASSP)}, 
  title={{Learning Gradients of Convex Functions with Monotone Gradient Networks}}, 
  year={2023},
  volume={},
  number={},
  pages={1-5},
  doi={10.1109/ICASSP49357.2023.10097266}}

@article{goujon2023neural,
  title={{A Neural-Network-based Convex Regularizer for Inverse Problems}},
  author={Goujon, Alexis and Neumayer, Sebastian and Bohra, Pakshal and Ducotterd, Stanislas and Unser, Michael},
  journal={IEEE Transactions on Computational Imaging},
  volume={9},
  pages={781--795},
  year={2023},
  publisher={IEEE}
}

@article{brenier1991polar,
  title={{Polar Factorization and Monotone Rearrangement of Vector-valued Functions}},
  author={Brenier, Yann},
  journal={Communications on Pure and Applied Mathematics},
  volume={44},
  number={4},
  pages={375--417},
  year={1991},
  publisher={Wiley Online Library}
}

@inproceedings{mukherjee2024data,
  title={{Data-driven Convex Regularizers for Inverse Problems}},
  author={Mukherjee, Subhadip and Dittmer, S{\"o}ren and Shumaylov, Zakhar and Lunz, Sebastian and {\"O}ktem, Ozan and Sch{\"o}nlieb, C-B},
  booktitle={ICASSP 2024-2024 IEEE International Conference on Acoustics, Speech and Signal Processing (ICASSP)},
  pages={13386--13390},
  year={2024},
  organization={IEEE}
}

@inproceedings{ConvexPotentialFlows,
  title={{Convex Potential Flows: Universal Probability Distributions with Optimal Transport and Convex Optimization}},
  author={Huang, Chin-Wei and Chen, Ricky TQ and Tsirigotis, Christos and Courville, Aaron},
  booktitle={International Conference on Learning Representations},
  year={2020}
}

@article{orlando2022unified,
  title={{A Unified Theory of Adaptive Subspace Detection part I: Detector Designs}},
  author={Orlando, Danilo and Ricci, Giuseppe and Scharf, Louis L.},
  journal={IEEE Transactions on Signal Processing},
  volume={70},
  pages={4925--4938},
  year={2022},
  publisher={IEEE}
}

@book{rockafellar1970convex,
  title={{Convex Analysis}},
  author={Rockafellar, R. Tyrrell},
  volume={18},
  year={1970},
  publisher={Princeton University Press}
}

@inproceedings{
AMIPS,
title={{Amortizing Maximum Inner Product Search with Learned Support Functions}},
author={Theo X. Olausson and Joao Monteiro and Michal Klein and Marco Cuturi},
booktitle={Forty-third International Conference on Machine Learning},
year={2026},
url={https://openreview.net/forum?id=0TRMWCwKRo}
}

@inproceedings{ICNN,
  title={{Input Convex Neural Networks}},
  author={Amos, Brandon and Xu, Lei and Kolter, J. Zico},
  booktitle={International Conference on Machine Learning},
  pages={146--155},
  year={2017},
  organization={PMLR}
}

@article{ICGN,
  title={{Input Convex Gradient Networks}},
  author={Richter-Powell, Jack and Lorraine, Jonathan and Amos, Brandon},
  journal={arXiv preprint arXiv:2111.12187},
  year={2021}
}

@INPROCEEDINGS{CAMSAPGradNets,
  author={Chaudhari, Shreyas and Pranav, Srinivasa and Moura, José M. F.},
  booktitle={2025 IEEE 10th International Workshop on Computational Advances in Multi-Sensor Adaptive Processing (CAMSAP)}, 
  title={{GradNetOT: Learning Optimal Transport Maps with GradNets}}, 
  year={2025},
  volume={},
  number={},
  pages={326-330},
  doi={10.1109/CAMSAP66162.2025.11423958}}

@inproceedings{GlobalOptimality,
  title={{Global Optimality in Neural Network Training}},
  author={Haeffele, Benjamin D and Vidal, Ren{\'e}},
  booktitle={2017 IEEE Conference on Computer Vision and Pattern Recognition (CVPR)},
  pages={4390--4398},
  year={2017},
  organization={IEEE}
}

@ARTICLE{MatchedSubspaceDetector,
  author={Scharf, L.L. and Friedlander, B.},
  journal={IEEE Transactions on Signal Processing}, 
  title={{Matched Subspace Detectors}}, 
  year={1994},
  volume={42},
  number={8},
  pages={2146-2157},
  doi={10.1109/78.301849}}

@ARTICLE{BeamForming,
  author={Feldman, D.D. and Griffiths, L.J.},
  journal={IEEE Transactions on Signal Processing}, 
  title={{A Projection Approach for Robust Adaptive Beamforming}}, 
  year={1994},
  volume={42},
  number={4},
  pages={867-876},
  doi={10.1109/78.285650}}

@inproceedings{AsymmetricALSH,
 author = {Shrivastava, Anshumali and Li, Ping},
 booktitle = {Advances in Neural Information Processing Systems},
 pages = {},
 publisher = {Curran Associates, Inc.},
 title = {{Asymmetric LSH (ALSH) for Sublinear Time Maximum Inner Product Search (MIPS)}},
 url = {https://proceedings.neurips.cc/paper_files/paper/2014/file/c98e7c4b8f20d384e3ad857d0ee226cc-Paper.pdf},
 volume = {27},
 year = {2014}
}

@inproceedings{TreeCones,
author = {Ram, Parikshit and Gray, Alexander G.},
title = {{Maximum Inner-product Search using Cone Trees}},
year = {2012},
isbn = {9781450314626},
publisher = {Association for Computing Machinery},
address = {New York, NY, USA},
url = {https://doi.org/10.1145/2339530.2339677},
doi = {10.1145/2339530.2339677},
booktitle = {Proceedings of the 18th ACM SIGKDD International Conference on Knowledge Discovery and Data Mining},
pages = {931–939},
numpages = {9},
location = {Beijing, China},
series = {KDD '12}
}

@article{jihong2026monotone,
  title={{Monotone Peridynamic Neural Operator for Nonlinear Material Modeling with Conditionally Unique Solutions}},
  author={Wang, Jihong and Tian, Xiaochuan and Zhang, Zhongqiang and Silling, Stewart and Jafarzadeh, Siavash and Yu, Yue},
  journal={Computer methods in applied mechanics and engineering},
  volume={453},
  pages={118792},
  year={2026},
  publisher={Elsevier}
}

@article{jacob2025generalizable,
  title={{Generalizable Physics-informed Learning for Stochastic Safety-critical Systems}},
  author={Wang, Zhuoyuan and Chern, Albert and Nakahira, Yorie},
  journal={IEEE Transactions on Automatic Control},
  year={2025},
  publisher={IEEE}
}

@phdthesis{noufal2026generative,
  title={{On Generative Flow Models for Stochastic Dynamical Systems: Paths and Densities}},
  author={El Bekri, Naoufal},
  year={2026},
  school={Universit{\'e} de Brest}
}

@article{Hinkkanen2026gradient,
  title={{Gradient Networks for Universal Magnetic Modeling of Synchronous Machines}},
  author={Li, Junyi and Foissner, Tim and Martin, Floran and Piippo, Antti and Hinkkanen, Marko},
  journal={arXiv preprint arXiv:2602.14947},
  year={2026}
}

@article{benning2018modern,
  title={{Modern Regularization Methods for Inverse Problems}},
  author={Benning, Martin and Burger, Martin},
  journal={Acta numerica},
  volume={27},
  pages={1--111},
  year={2018},
  publisher={Cambridge University Press}
}

\end{document}